\documentclass[conference,compsoc]{IEEEtran}
\ifCLASSOPTIONcompsoc
  \usepackage[nocompress]{cite}
\else
  \usepackage{cite}
\fi

\usepackage{amsmath,amssymb}
\usepackage{algorithmic}

\usepackage{multirow}
\usepackage{tikz,pgfplots}
\usetikzlibrary{matrix,chains,positioning,decorations.pathreplacing,arrows}
\usepackage{adjustbox}
\usepackage[tight,footnotesize]{subfigure}
\usepackage{graphicx,algorithm}
\DeclareMathOperator*{\argmin}{argmin}
\begin{document}

\title{Robust Local Scaling using Conditional Quantiles of Graph Similarities}

\author{\IEEEauthorblockN{Jayaraman J. Thiagarajan}
\IEEEauthorblockA{Lawrence Livermore National Labs\\
7000 E Avenue Livermore, CA 94550\\
Email: jjayaram@llnl.gov}
\and
\IEEEauthorblockN{Prasanna Sattigeri, Karthikeyan \\Natesan Ramamurthy}
\IEEEauthorblockA{IBM Research, Yorktown Heights, NY 10568\\
Email: \{psattig, knatesa\}@us.ibm.com}
\and
\IEEEauthorblockN{Bhavya Kailkhura}
\IEEEauthorblockA{Syracuse University\\
Syracuse, NY 13244\\
Email: bkailkhu@syr.edu}}



%

\maketitle

\begin{abstract}
Spectral analysis of neighborhood graphs is one of the most widely used techniques for exploratory data analysis, with applications ranging from machine learning to social sciences. In such applications, it is typical to first encode relationships between the data samples using an appropriate similarity function. Popular neighborhood construction techniques such as k-nearest neighbor (k-NN) graphs are known to be very sensitive to the choice of parameters, and more importantly susceptible to noise and varying densities. In this paper, we propose the use of quantile analysis to obtain local scale estimates for neighborhood graph construction. To this end, we build an auto-encoding neural network approach for inferring conditional quantiles of a similarity function, which are subsequently used to obtain robust estimates of the local scales. In addition to being highly resilient to noise or outlying data, the proposed approach does not require extensive parameter tuning unlike several existing methods. Using applications in spectral clustering and single-example label propagation, we show that the proposed neighborhood graphs outperform existing locally scaled graph construction approaches. 
\end{abstract}


\IEEEpeerreviewmaketitle

\section{Introduction}
Neighborhood graphs are central to techniques that involve analysis and exploration of high-dimensional data. Constructing a graph involves encoding the relationships between the data samples using an appropriate similarity function. Spectral analysis of such graphs is the modus operandi in a variety of applications, including dimensionality reduction, image segmentation, text mining, and data analysis in general. These methods often involve the eigen-decomposition of the similarity (also referred to as the adjacency) matrix that reveal strong connections to graph properties such as connected components, the diameter of a graph and the degree of randomness.

Defining the notion of an appropriate neighborhood and adapting the analysis to the local scale or density of the data have been long-standing research problems. $k-$nearest neighborhood or $\epsilon-$neighborhood graphs are the most commonly adopted approaches in practice. However, the instabilities arising due to noise or outlying data have plagued the performance of nearest neighbor graphs. Furthermore, clusters with varying densities commonly occur in high-dimensions, which make the global neighborhood parameter choices highly unreliable. Consequently, a broad class of techniques that attempt to estimate the local scale to improve spectral analysis of data with varying densities, shapes and noise have been developed \cite{perona}. Another class of approaches translate the density variations into edge probabilities representing their significance in recovering the underlying structure \cite{race2014flexible}. Several of these methods still rely heavily on heuristics and parameter tuning to perform consistently across different domains. 

In this paper, we explore the use of quantile analysis to obtain local scale estimates for building robust similarities. The proposed approach falls under the class of methods that construct stochastic graphs, which is carried out based on a novel, unsupervised quantile analysis framework. Quantile regression has been primarily used in analysis of  datasets with heterogeneous properties, wherein traditional loss functions such as the $\ell_2$ fail to account for biases in different parts of the data. In our context, spectral decomposition of graphs using squared $\ell_2$ loss is widely used with a stable numerical solution (eigen decomposition). However, when we use large graphs and wish to discover its non-linear spectral structure at various quantiles, it becomes imperative to use a general optimization procedure. To this end, we build an auto-encoding neural network, which imposes a quantile loss between the input and reconstructed similarity matrices, to infer the conditional quantiles of similarity functions on graphs. The conditional quantiles are subsequently used to obtain robust estimates of the local scale. We show that the resulting neighborhood graphs outperform existing locally scaled graph construction approaches. Furthermore, we demonstrate through our experiments that the proposed method is highly resilient to noise and does not require extensive parameter tuning. Our contributions can be summarized as follows:
\begin{itemize}
\item We generalize the notion of quantile analysis to the similarity function defined on unsupervised graphs.
\item We build a neural network architecture for efficient inference of conditional quantiles of neighborhood similarities.
\item We relate the rate of edge decay across quantiles to the edge probabilities while constructing stochastic neighborhood graphs.
\item Using the proposed stochastic graphs, we develop a robust local scale estimation algorithm.
\item We demonstrate that the proposed approach is resilient to noise and variations in local densities, and consistently outperforms existing approaches for spectral clustering.
\item We propose a greedy algorithm for using the inferred graph similarities in label propagation with limited training examples.
\item In an extreme setting, we evaluate the robustness of the graphs in label propagation with a single example (per class) and demonstrate substantial improvements.
\end{itemize}

%


\section{Related Work}
Constructing neighborhood graphs and performing spectral analysis of pairwise affinities are common to a wide-range of applications dealing with complex, high-dimensional data. Simple neighborhood techniques such as k-nearest neighbor (k-NN) graphs are known to be very sensitive to the choice of parameters. Alternatively, one can simply connect all points in a fully connected graph and rely on a scaling parameter $\sigma$ to define the affinity between two points. In either case, the technique relies on setting a global parameter that does not take into account the variations in local densities. Consequently, several improvements have been proposed in the literature, through analysis of the underlying graph structure \cite{perona,MaiervH2012,Bicici2007}, and the stability \cite{Huang2009,vonluxburg2008} of spectral clustering. In particular, techniques that obtain estimates of the local scale to handle data with varying densities and levels of noise have gained significant interest. One of the earliest strategies to estimate the local scales was developed by Zelnik-Manor and Perona \cite{perona}. Though this approach is effective even in high dimensions, it often leads to sub-optimal performances in presence of outliers/noise and in data with clusters of different densities. To alleviate this, Nadler \textit{et al.} employed a coherence measure of a set of points for belonging to the same cluster \cite{nadler2007}. Furthermore, Li \textit{et al.} \cite{Li2007} proposed a warping model that maps the data into a new feature space for reliable clustering. Another interesting approach for local scale estimation stems from the use of proximity graphs (e.g. beta skeletons) to infer the neighborhood parameters \cite{Correa2012}, which was found to be resilient to noise.

Since pairwise similarities are solely based on the Euclidean distances between samples in the input space, they reveal no information about the inherent class structure. An effective approach to capture that information is to exploit the underlying manifold structure so that samples belonging to the same manifold have consistently higher similarity while samples belonging to different manifolds do not. The path-based graphs \cite{Chang2005} define a similarity measure that implicitly infers the underlying structure and produces a robust neighborhood graph. Another important class of approaches for graph construction attempt to build probabilistic graphs that reveal the relative significance of the different edges \cite{Potamias2010,meyer2012stochastic}. In particular, the consensus clustering algorithm in \cite{race2014flexible} circumvents the problem of parameter selection by creating an ensemble of clustering with different parameter choices and exploiting the theory of nearly uncoupled Markov chains to construct a probabilistic graph, which is finally used with spectral clustering for robust analysis. This iterative procedure can be practically infeasible in high-dimensions, and their performance can be affected by varying densities. In this paper, we propose to adopt ideas from quantile analysis to construct robust graph similarities and thereby alleviate the inherent challenges with local scale estimation algorithms.
  
\section{Inferring Conditional Quantiles of Graph Similarity}
Supervised regression is a common statistical approach employed to analyze the relationships between the predictor variables and a response variable. Regression with squared $\ell_2$ loss, \emph{aka} the method of least squares, estimates the conditional mean of the response variable for the given predictors. This is sufficient when the data is homogeneous; however, when the data is heterogeneous, merely estimating the conditional mean is insufficient, as estimates of the standard errors are often biased. To comprehensively analyze such heterogeneous datasets, quantile regression is a better alternative. Quantile regression aims at estimating either the conditional median or other quantiles of the response variable \cite{koenker1978regression, takeuchi2006nonparametric}.

\subsection{Definition: Quantile Loss}
Quantile losses as fidelity measures for function approximation has applications in computational biology \cite{Zou08}, survival analysis \cite{KG01}, workforce analytics \cite{Ramamurthy2013GlobalSIP}, economics \cite{KH01} and data analysis \cite{Bhatele2015, Ramamurthy2016} to name a few. For a scalar residual $r$, the quantile loss is a \emph{check function} defined as
\begin{equation}
\label{en:quant_loss}
q_{\tau}(r) = (-\tau+1 [{r \geq 0}]) r.
\end{equation} The quantile loss is piecewise linear and shares the robustness properties of the $\ell_1$ loss by not penalizing the outliers as harshly as the squared $\ell_2$ loss. Note that in (\ref{en:quant_loss}) the quantile loss is equivalent to the $\ell_1$ loss when $\tau=0.5$. Consequently, similar to $\ell_1$, the quantile loss is non-differentiable at the origin and forces the residuals close to the origin to be exactly zero which may not be preferred in some applications.

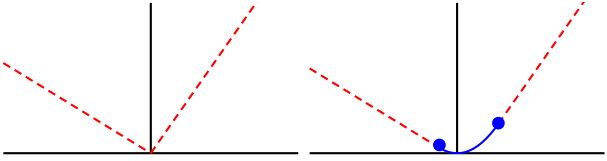
\begin{figure}[t]
\centering
\begin{minipage}{0.49\textwidth} 
\begin{tikzpicture}
  \begin{axis}[
    thick,
    width=.45\linewidth, height=2cm,
    xmin=-2,xmax=2,ymin=0,ymax=1,
    no markers,
    samples=50,
    axis lines*=left, 
    axis lines*=middle, 
    scale only axis,
    xtick={0},
    ytick={0},
    ] 
\addplot[red,domain=-2:0,densely dashed]{-.3*x};
\addplot[red,domain=0:+2,densely dashed]{.7*x};
  \end{axis}
\end{tikzpicture}
\begin{tikzpicture}
  \begin{axis}[
    thick,
    width=.45\linewidth, height=2cm,
    xmin=-2,xmax=2,ymin=0,ymax=1,
    no markers,
    samples=50,
    axis lines*=left, 
    axis lines*=middle, 
    scale only axis,
    xtick={0},
    ytick={0},
    ] 
\addplot[red,domain=-2:-2*0.3*0.4,densely dashed]{0.3*abs(x) - 0.4*0.3^2};
\addplot[blue,domain=-2*0.3*0.4:2*(1-0.3)*0.4]{0.25*x^2/0.4};
\addplot[red,domain=2*(1-0.3)*0.4:2,densely dashed]{(1-0.3)*abs(x) - 0.4*(1-0.3)^2};
\addplot[blue,mark=*,only marks] coordinates {(-.24,0.0550) (0.56,0.20)};
  \end{axis}
\end{tikzpicture}



\end{minipage}
\caption{Quantile $(\tau = 0.3)$  loss  (left) and quantile Huber $(\tau = 0.3)$ loss (right). 
The quantile Huber loss is obtained by smoothing the quantile loss at the origin.}
\label{fig:PLQFig}
\end{figure}

A smoothed `huberized' version of the quantile loss has been recently proposed \cite{aravkin2014qh}. This is defined as
\begin{equation}
\label{quantileHuber}
\small
\rho_\tau(r) = \begin{cases}
\tau |r| - \frac{\kappa \tau^2}{2} & \text{if} \text{ } r < -\tau\kappa,\\
\frac{1}{2\kappa}r^2 & \text{if} \text{ } r\in [-\kappa \tau, (1-\tau)\kappa],\\ 
(1-\tau) |r| - \frac{\kappa(1-\tau)^2}{2}, & \text{if} \text{ } r > \quad (1-\tau)\kappa.
\end{cases} 
\end{equation} The quantile Huber loss permits small residuals by behaving like a squared $\ell_2$ loss near the origin and hence may be preferred in regression settings. Figure \ref{fig:PLQFig} shows the quantile $(\tau = 0.3)$ and quantile Huber $(\tau = 0.3)$ loss functions. Furthermore, the quantile loss itself is a special case of quantile Huber as $\kappa \rightarrow 0$. Both quantile and quantile Huber losses are additive along the co-ordinates of the residual and hence the loss for multi-dimensional residuals is easily defined as $\rho_\tau(\mathbf{r}) = \sum_{i=1}^N \rho_\tau (r_i)$. In the rest of the paper, we will use the Huberized version of quantile loss.

\subsection{Quantile Huber Loss in Spectral Graph\\ Decomposition}
\label{sec:qloss}
Although quantile losses are traditionally used only in regression settings, it can be beneficial to employ them in unsupervised learning where the goal is to explore the structure of the data in lieu of fitting a function. Alternately, we could consider the response variable to be as the input data itself and thereby infer the underlying structure using the loss. However, the meaning of this formulation is problem-specifc and needs careful consideration. We will focus on the problem of spectral decomposition of graphs in this paper.

Let us define a undirected graph with $N$ nodes denoted by its similarity matrix $\mathbf{W} \in \mathbb{R}^{N \times N}$. The $(i,j)^{\text{th}}$ element $w_{ij}$ corresponds to the similarity of the edge between the two nodes. $\mathbf{W}$ is symmetric positive semi-definite, has all positive entries, and the maximum value of the entries is $1$. The low-rank decomposition of this graph can be used for applications such as spectral clustering. For simplicity, we will consider an L-R decomposition of $\mathbf{W}$, where $\mathbf{L} \in \mathbb{R}^{N \times P}$ and  $\mathbf{R} \in \mathbb{R}^{N \times P}$. In this case, our goal is to measure the fidelity of the approximation $\mathbf{\hat{W}} = \mathbf{L} \mathbf{R}^T$. The corresponding optimization with quantile loss is posed as
\begin{equation}
\min_{\mathbf{L}, \mathbf{R}} \rho_{\tau} (\mathbf{W}-\mathbf{L} \mathbf{R}^T).
\label{eqn:LR}
\end{equation}Setting $P=1$ for simplicity, this becomes
\begin{equation}
\min_{\mathbf{l}, \mathbf{r}}  \sum_{i,j}\rho_{\tau}(w_{ij}- l_i r_j).
\label{eqn:LR_1}
\end{equation}When $\tau$ is high, the positive residuals will be penalized less compared to the negative residuals and hence $l_i r_j$ will underestimate $w_{ij}$ for most $i$ and $j$. Setting the negative values of  $l_i r_j $ to zero will lead to a sparse similarity matrix. On the other hand, lower values of $\tau$ will correspond to lesser number of negative values in $\{l_i r_j | \forall i, j \in 1, \ldots, N\}$, thereby resulting in a dense graph. Hence, we can consider the spectral decomposition of graphs with quantile penalties as a way of robustly sparsifying the graph while preserving the essential spectral structure. Note, optimizing for $\hat{\mathbf{W}}$ without imposing any spectral structural constraints ($\hat{\mathbf{W}} = \mathbf{L} \mathbf{R}^T$) will result in all the elements of $\hat{\mathbf{W}}$ being equal to the $\tau^{\text{th}}$ quantile of the elements of $\mathbf{W}$. Hence it is important to preserve the spectral structure when obtaining the conditional quantile estimates.

\begin{figure}
\begin{center}
\begin{adjustbox}{max width=0.4\textwidth}
\begin{tikzpicture}[
plain/.style={
  draw=none,
  fill=none,
  },
net/.style={
  matrix of nodes,
  nodes={
    draw,
    circle,
    inner sep=10pt
    },
  nodes in empty cells,
  column sep=2cm,
  row sep=-9pt
  },
>=latex
]
\matrix[net] (mat)
{
|[plain]| \parbox{1.3cm}{\centering Input\\layer} & |[plain]| \parbox{1.3cm}{\centering Hidden\\layer} & |[plain]| \parbox{1.3cm}{\centering Output\\layer} \\
& |[plain]| & \\
|[plain]| & \\
& |[plain]| & \\
|[plain]| & |[plain]| \\
& & \\
|[plain]| & |[plain]| \\
& |[plain]| & \\
|[plain]| & \\
& |[plain]| & \\
};
\foreach \ai [count=\mi ]in {2,4,...,10}
  \draw[<-] (mat-\ai-1) -- node[above] {$x_\mi$} +(-2cm,0);
\foreach \ai in {2,4,...,10}
{\foreach \aii in {3,6,9}
  \draw[->] (mat-\ai-1) -- (mat-\aii-2);
}
\foreach \ai [count=\mi ]in {2,4,...,10}
  \draw[->] (mat-\ai-3) -- node[above] {$\hat{x}_\mi$} +(2cm,0);
\foreach \ai in {3,6,9}
{\foreach \aii in {2,4,...,10}
  \draw[->] (mat-\ai-2) -- (mat-\aii-3);
}

\end{tikzpicture}
\end{adjustbox}
\end{center}
\caption{An example architecture with $5-$dimensional inputs and $3-$dimensional representations in the hidden layer.}
\label{fig:AE}
\end{figure}
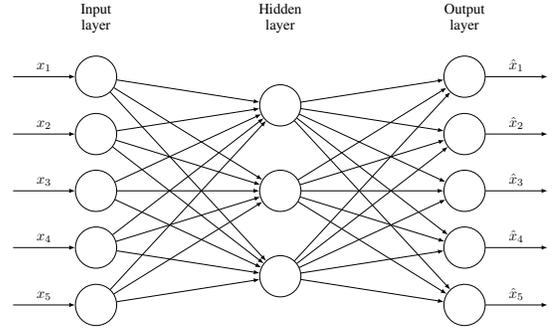

\subsection{Non-linear Spectral Decomposition using Auto-Encoders}
We propose a non-linear extension to the spectral decomposition approach described in the previous section. To infer conditional quantiles of graph similarity, we solve the problem $\rho_{\tau}(\mathbf{W} - \hat{\mathbf{W}}) $ using an auto-encoding neural network. An auto-encoder learns to reconstruct/predict its input by learning successive non-linear encodings and subsequent non-linear decodings using \emph{hidden layers}. Even though the network is learning an identity function, by placing constraints such as low-dimensionality, sparsity, etc. on the hidden layers, useful structure learning can be performed. The hidden unit activations can then be used as feature representations. In the simplest form, we can express the reconstructed similarity as the composition, $\hat{\mathbf{W}} = g\left(f\left(\mathbf{W}\right)\right)$, where $f$ is a non-linear transformation and $g$ is its inverse. The auto-encoder attempts to learn both $f$ and  $g$ by restricting them to specific forms of non-linearity. In particular the following forms are widely used and well-understood: $\mathbf{H} = f(\mathbf{W}) = \phi\left(\Psi^{T}\mathbf{W}\right)$ and $\hat{\mathbf{W}} = g(\mathbf{H}) = \phi\left(\Psi \mathbf{H}\right)$ where $\Psi \in \mathbb{R}^{N \times P}$ with $P < N$ and $\phi$ is an elementwise non-linearity (\textit{aka} the activation function), such as the logistic sigmoid function. Comparing this formulation to the linear decomposition $\hat{\mathbf{W}}  = \mathbf{L} \mathbf{R}^T$, one can readily draw parallels: $\mathbf{R}^T$ is the forward transformation $f$ and $\mathbf{L}$ is the inverse $g$. We can now denote the non-linear spectral learning with squared $\ell_2$ loss using the objective
\begin{align} 
\hat{\Psi}  = & \argmin_{\Psi } \left\| \mathbf{W}- \phi\left(\Psi \phi  \left( \Psi^{T}\mathbf{W}\right)\right) \right\| _{2}^{2}.
\label{eqn:AE_loss_1}
\end{align}This can be optimized using an auto-encoder where $\mathbf{W}$ input data and $\mathbf{\Psi}$ are the weights between the input and hidden layers. A similar framework has been considered by Tian \textit{et al.} \cite{Tian2014}, where the authors show the equivalence between spectral clustering and the objective of an auto-encoder. To learn conditional quantile estimates of the input similarity matrix, the optimization can be re-posed using the quantile Huber loss $\rho_\tau$ as,
\begin{align} 
\nonumber \hat{\Psi}  = & \argmin_{\Psi } \rho_{\tau}  \left(\mathbf{W}- \phi \left(\Psi \phi  \left(\Psi^{T}\mathbf{W} \right) \right) \right).
\label{eqn:AE_loss_2}
\end{align} An example architecture is provided in Figure \ref{fig:AE} with one hidden layer. This can be generalized easily to multiple hidden layers resulting in deep and more complex representations. Interestingly, the linear L-R decomposition can be recovered by considering a single hidden layer with $P$ units ($P < N$), and $\phi$ set to the identity function. As a result, $\Psi^T \mathbf{W} = \mathbf{R}^T$ and $\Psi  = \mathbf{L}$. The ease of back-propagation based optimization enables the incorporation of additional constraints on the hidden layer weights and activation. In fact, this is equivalent to adding regularization terms on $\mathbf{R}$ and $\mathbf{L}$ in the linear case. 


We implement the above approach using stochastic gradient descent with mini-batch operations and automatic differentiation \cite{autograd}. The derivative of the quantile Huber loss with respect to the residual $r$ can be obtained as
\begin{equation}
\label{quantileHuber}
\small
\rho'_\tau(r) = \begin{cases}
-\tau & \text{if} \text{ } r < -\tau\kappa,\\
\frac{r}{\kappa} & \text{if} \text{ } r\in [-\kappa \tau, (1-\tau)\kappa],\\ 
(1-\tau) & \text{if} \text{ } r > \quad (1-\tau)\kappa.
\end{cases} 
\end{equation} Based on this, we create a custom automatic differentiation procedure for quantile Huber so that the gradients can be back-propagated through the network. 
%
%
\label{sec:optimization}

\section{Proposed Local Scale Estimation}
We propose to use the conditional quantiles of the graph similarity function to obtain estimates of the local scale for each sample in the dataset. To this end, we first construct a stochastic graph by studying the persistence of edges in reconstructed graphs at different quantiles, and then create random realizations of the neighborhood to obtain a robust estimate of the scale parameter.

\subsection{Constructing Stochastic Graphs}
The accuracy of spectral clustering depends, among other factors, on the appropriate choice of the scale parameter (and $k$ in $k-$NN graphs). Since global neighborhood parameters are insufficient for modeling disparate sampling densities across different clusters, it is common to define a more general, dense affinity matrix that incorporates local scaling. A popular alternative to choosing a single parameter is to define the affinity between two samples $\mathbf{x}_i$ and $\mathbf{x}_j$ as follows:
\begin{equation}
w_{ij} = \exp\left(-\frac{d(\mathbf{x}_i,\mathbf{x}_j)}{\sigma_i \sigma_j}\right),
\label{eqn:localscaling}
\end{equation}where $d(.,.)$ denotes an appropriate distance function and $\sigma_i$, $\sigma_j$ are the local scales corresponding to the samples $\mathbf{x}_i$ and $\mathbf{x}_j$ respectively. For example, in \cite{perona}, this parameter is defined as $\sigma_i = d(\mathbf{x}_i, \mathbf{x}_k^i)$, where $\mathbf{x}_k^i$ is the $k^{\text{th}}$ nearest neighbor of $\mathbf{x}_i$. While this graph similarity construction tends to produce improved results in practice, it still relies heavily on the choice of the parameter $k$ (set to $7$ in \cite{perona}). Furthermore, a single value of $k$ may not cluster data effectively in the presence of noise or under non-linear geometric transformations \cite{Correa2012}. In this paper, we argue that exploring the conditional quantiles of graph similarities will enable us to obtain good estimates of local scale thereby leading to robust performances.

\begin{algorithm}[t]
  \label{alg:scaleest}
  
  \begin{enumerate}
  \item Compute the locally scaled affinity matrix, $\mathbf{W}_{\ell}$, using (\ref{eqn:localscaling})
  \end{enumerate}
  \noindent \textbf{Construct stochastic graph:}
  \begin{enumerate}
  \setcounter{enumi}{1}
  \item Train autoencoders with the quantile huber loss for $\mathbf{W}_{\ell}$ (Section \ref{sec:optimization}), at different values of the quantile parameter $\tau$
  \item Compute the number of edges, $\beta_{\tau}$, at each quantile
  \item Construct a stochastic graph with edge probabilities given in (\ref{eqn:prob})
  \end{enumerate}
  \noindent \textbf{Estimate local scale:}
  \begin{enumerate}
  \setcounter{enumi}{4}
  \item Draw $R$ independent realizations of the neighborhood graph from the edge probabilities
  \item For each neighborhood graph, estimate the local scale of a sample as the average distance to all its neighbors
  \item Obtain the final local scale estimates as the median of the $R$ realizations
  \end{enumerate}
  \caption{Estimate the local scale for all samples in the input data $\mathbf{X}$} 
  \end{algorithm}
  
  \begin{figure*}[t]
\centering
  \subfigure[Input Data]{\includegraphics[width=.23\linewidth]{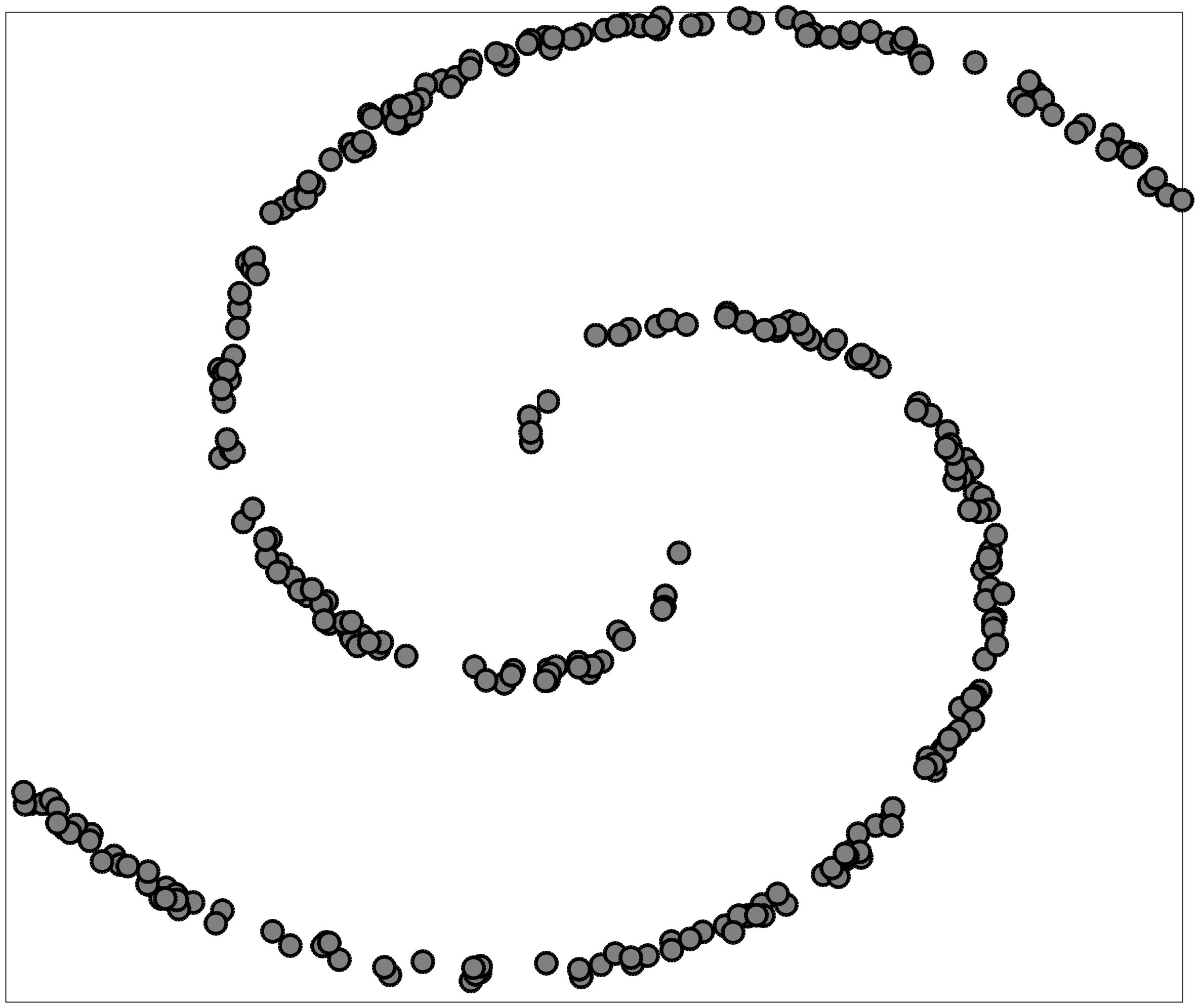}}\hfill
  \subfigure[$\tau=0.1$]{\includegraphics[width=.23\linewidth]{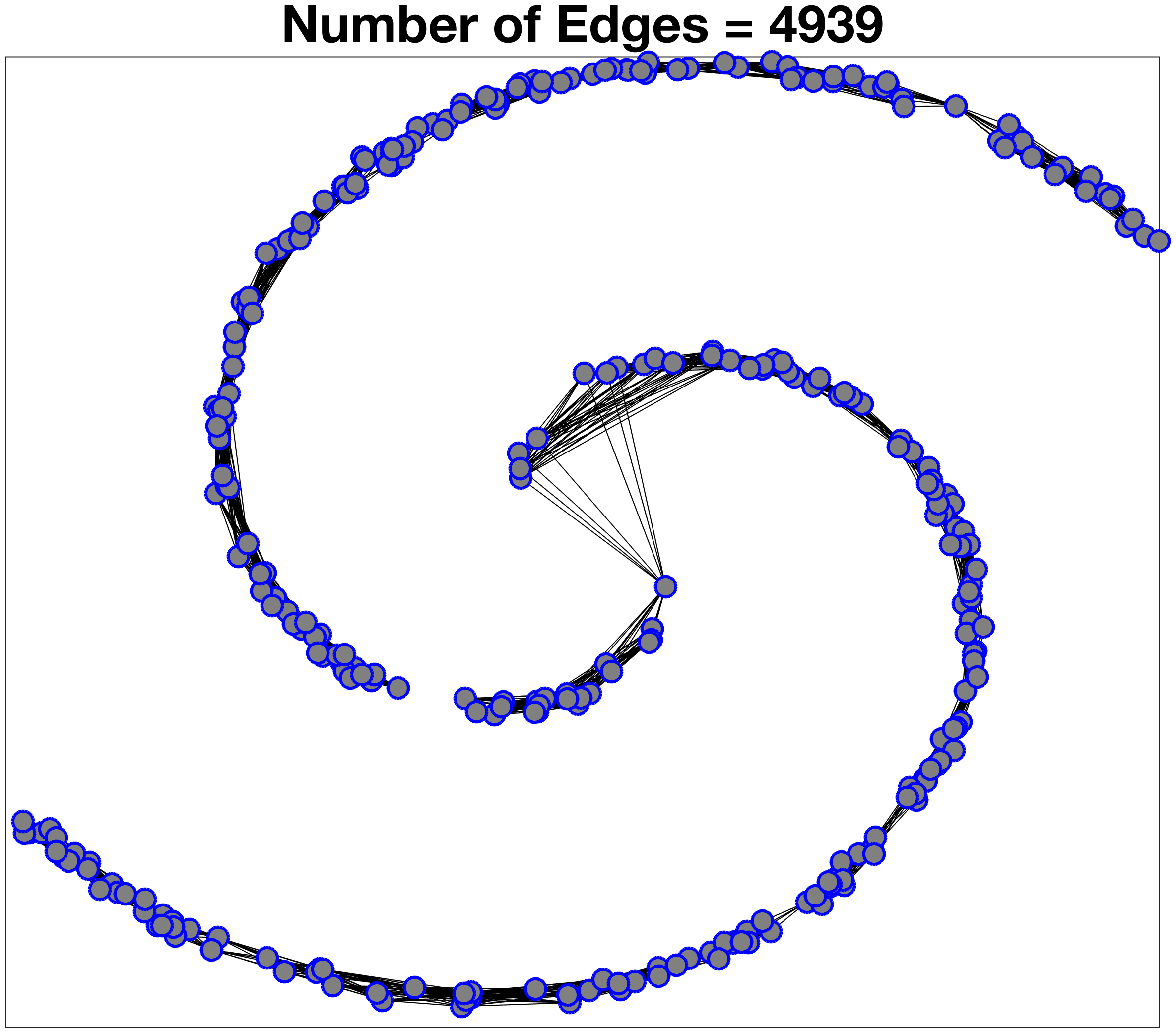}}\hfill
  \subfigure[$\tau=0.3$]{\includegraphics[width=.23\linewidth]{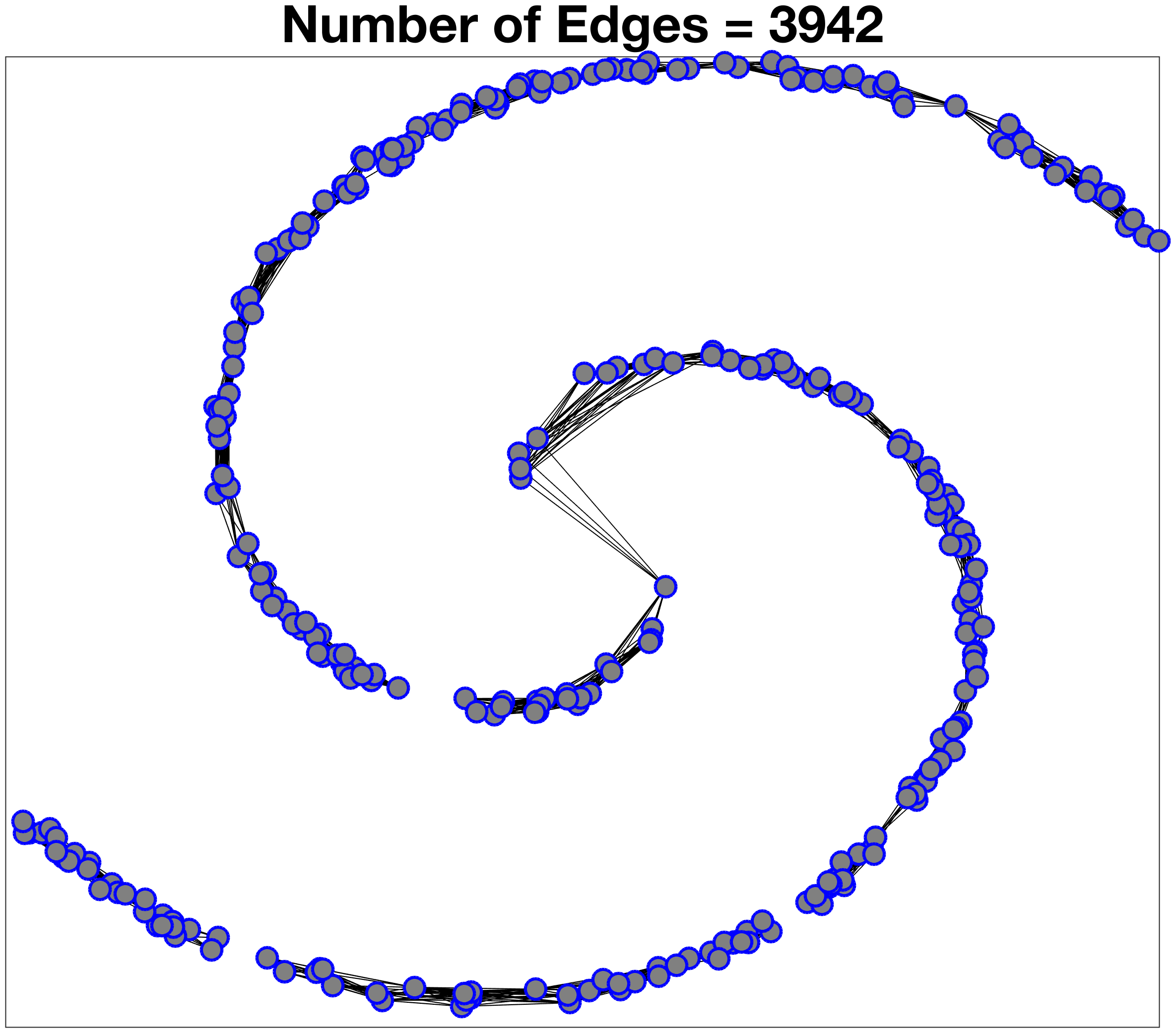}}\hfill
  \subfigure[$\tau=0.7$]{\includegraphics[width=.23\linewidth]{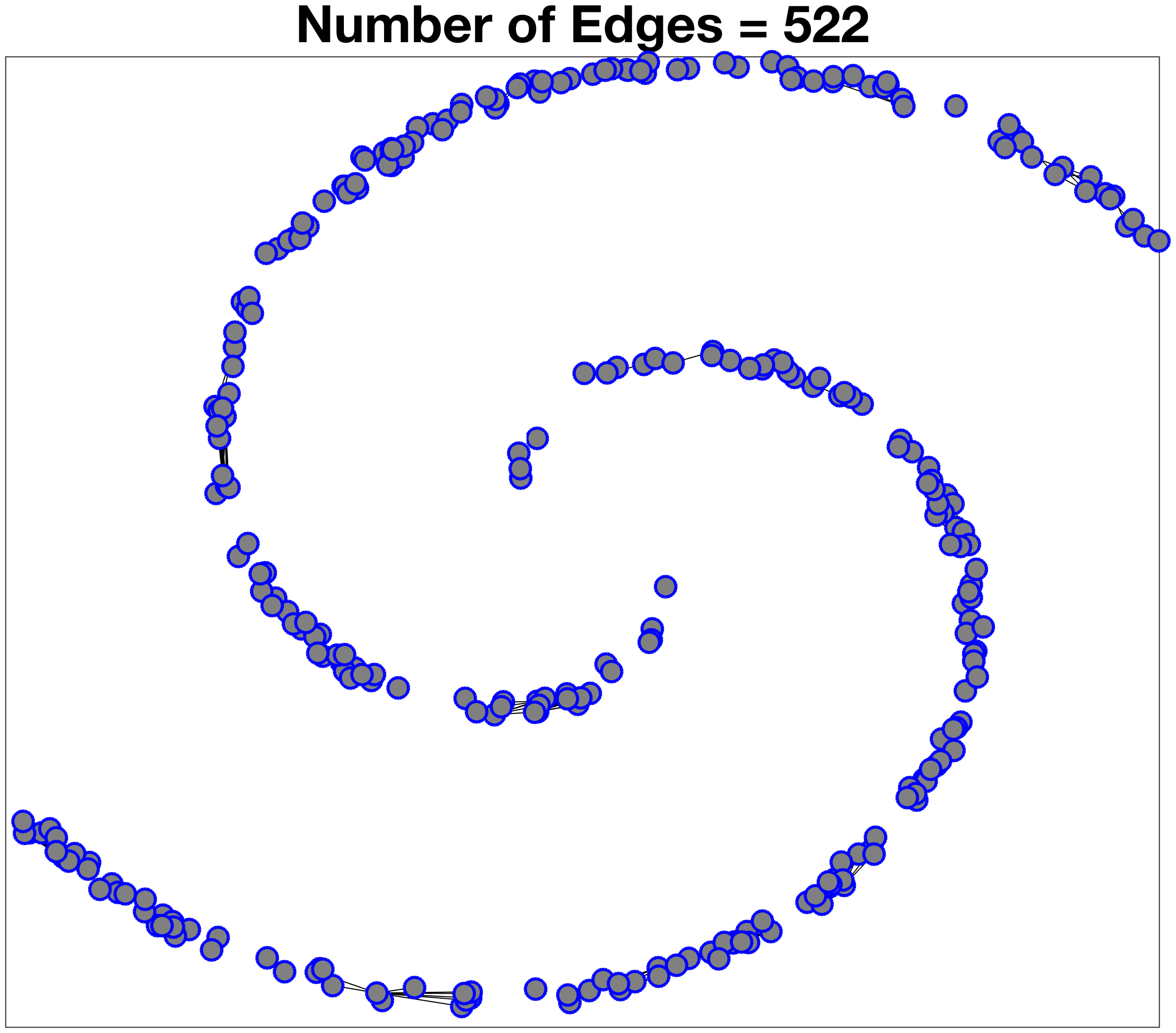}}
  \vfill
   \subfigure[Edge Decay]{\includegraphics[width=.23\linewidth]{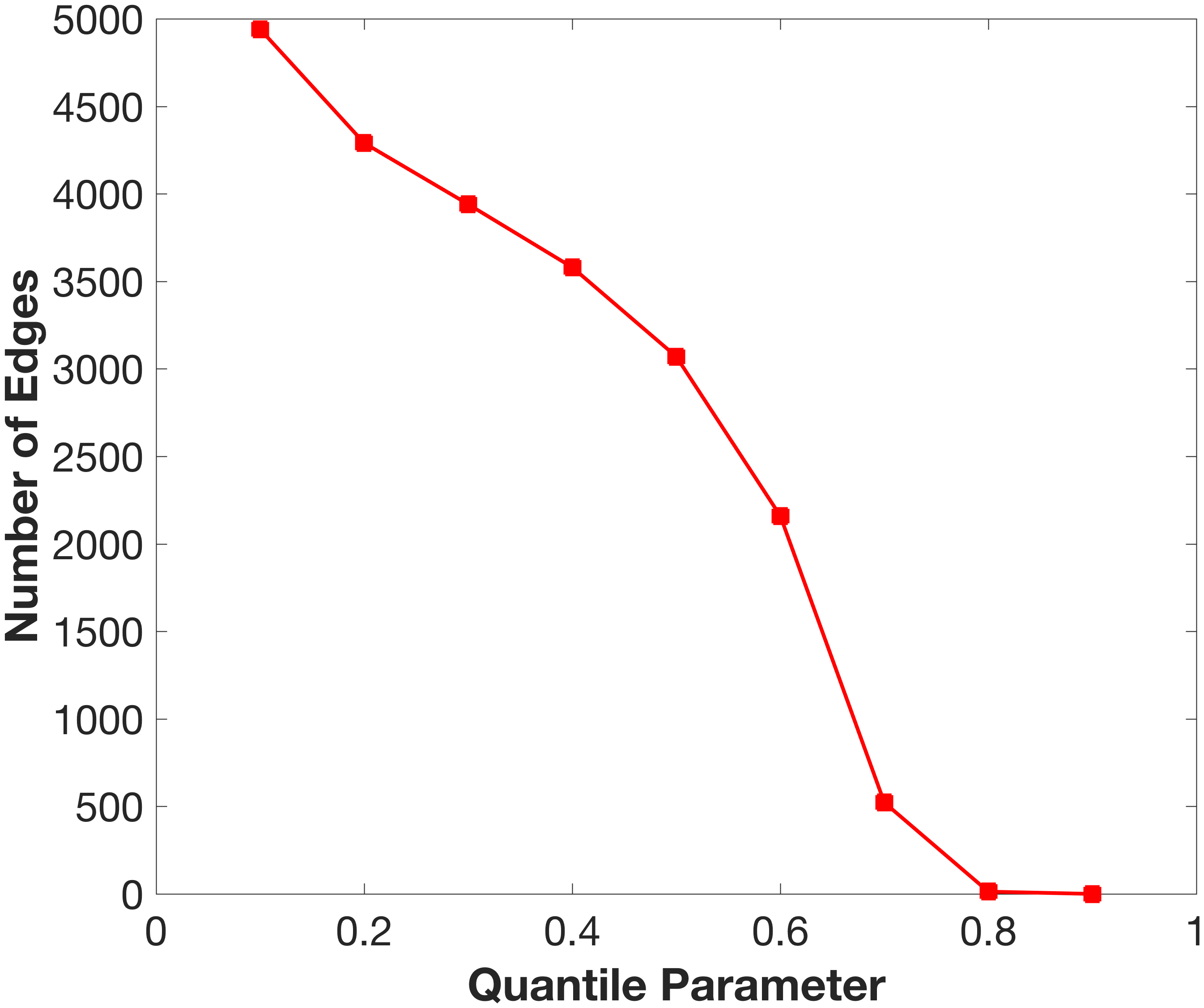}}\hfill
  \subfigure[Stochastic Graph]{\includegraphics[width=.23\linewidth]{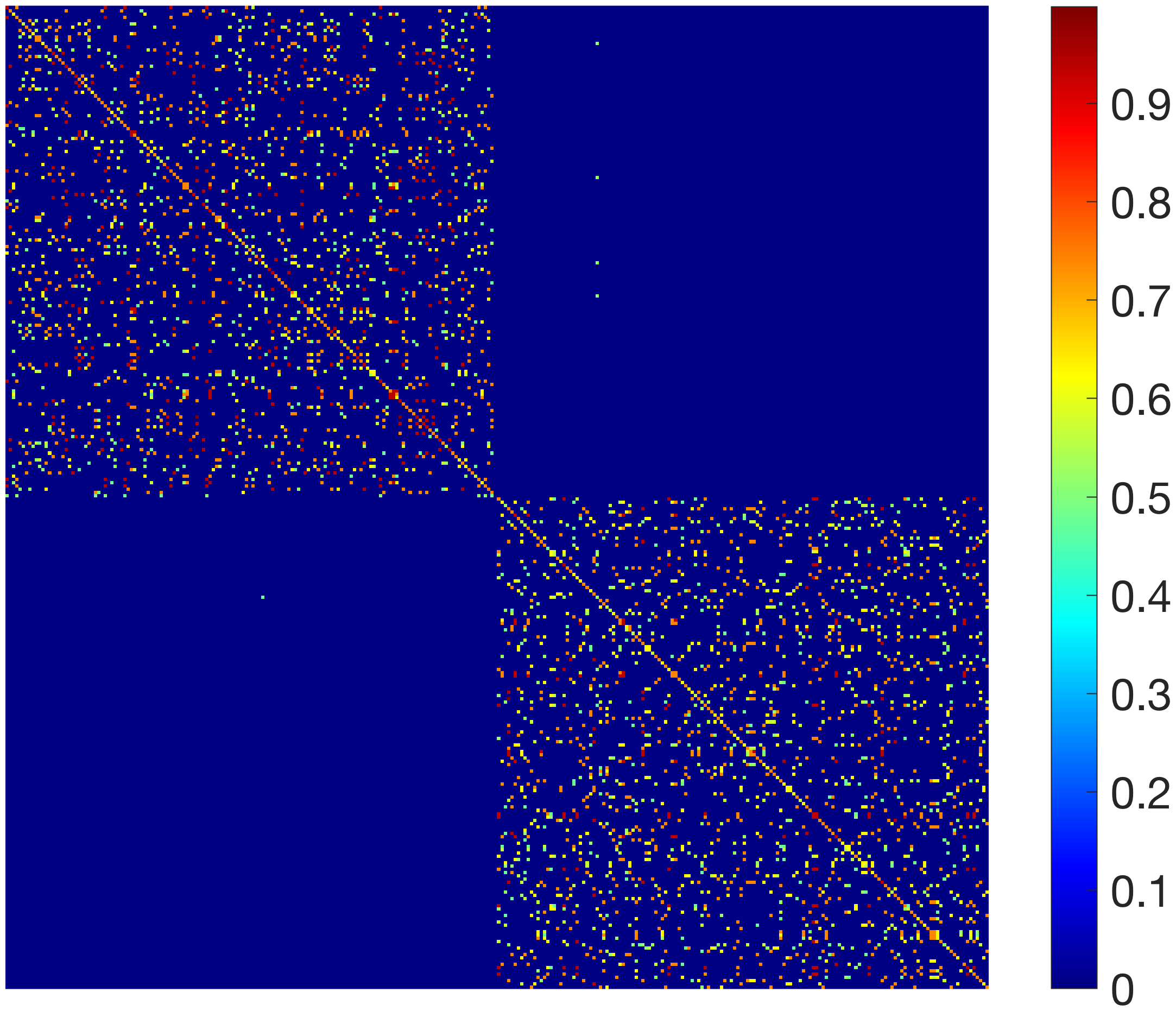}}\hfill
  \subfigure[Local Scaling]{\includegraphics[width=.23\linewidth]{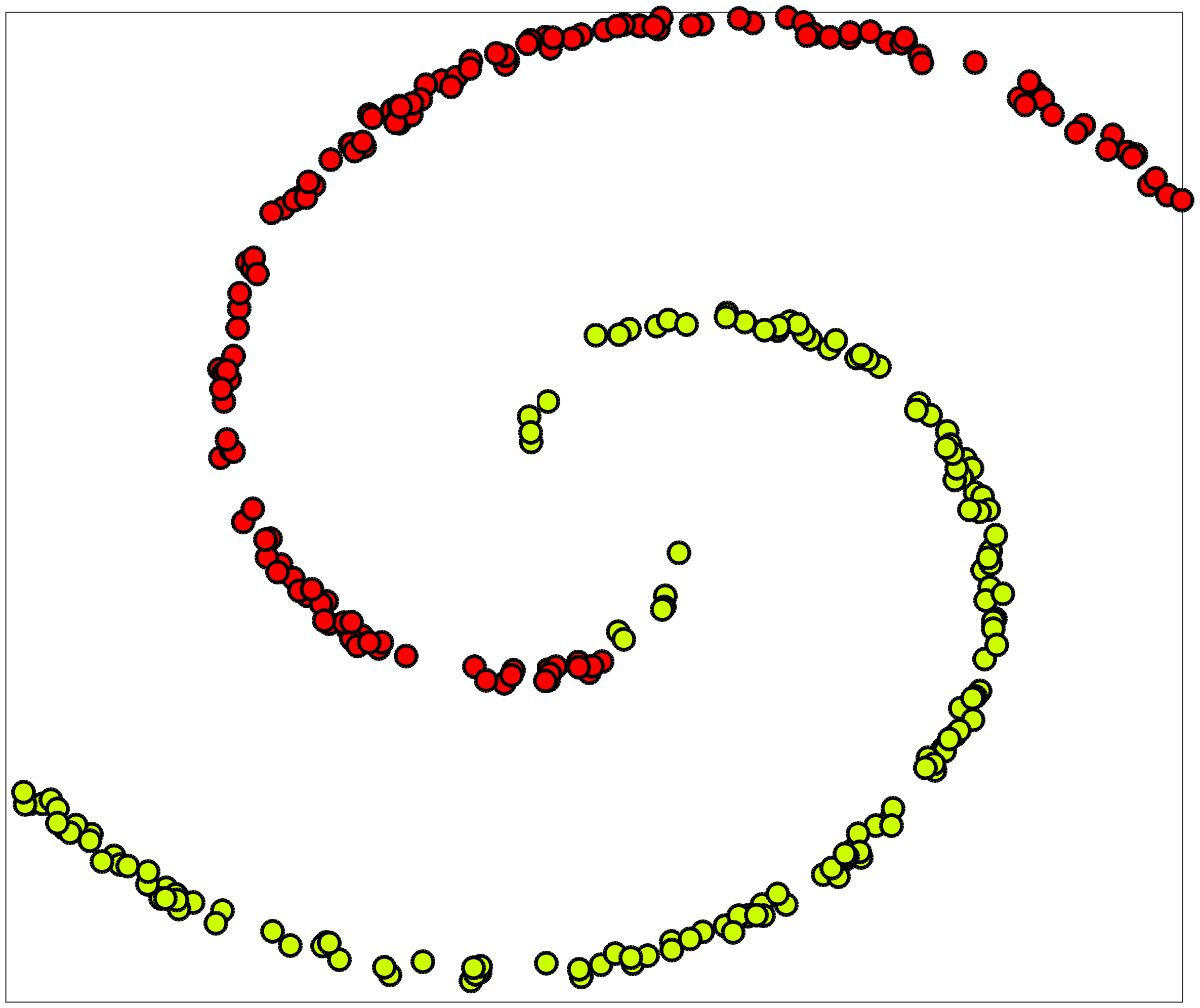}}\hfill
  \subfigure[Proposed Approach]{\includegraphics[width=.23\linewidth]{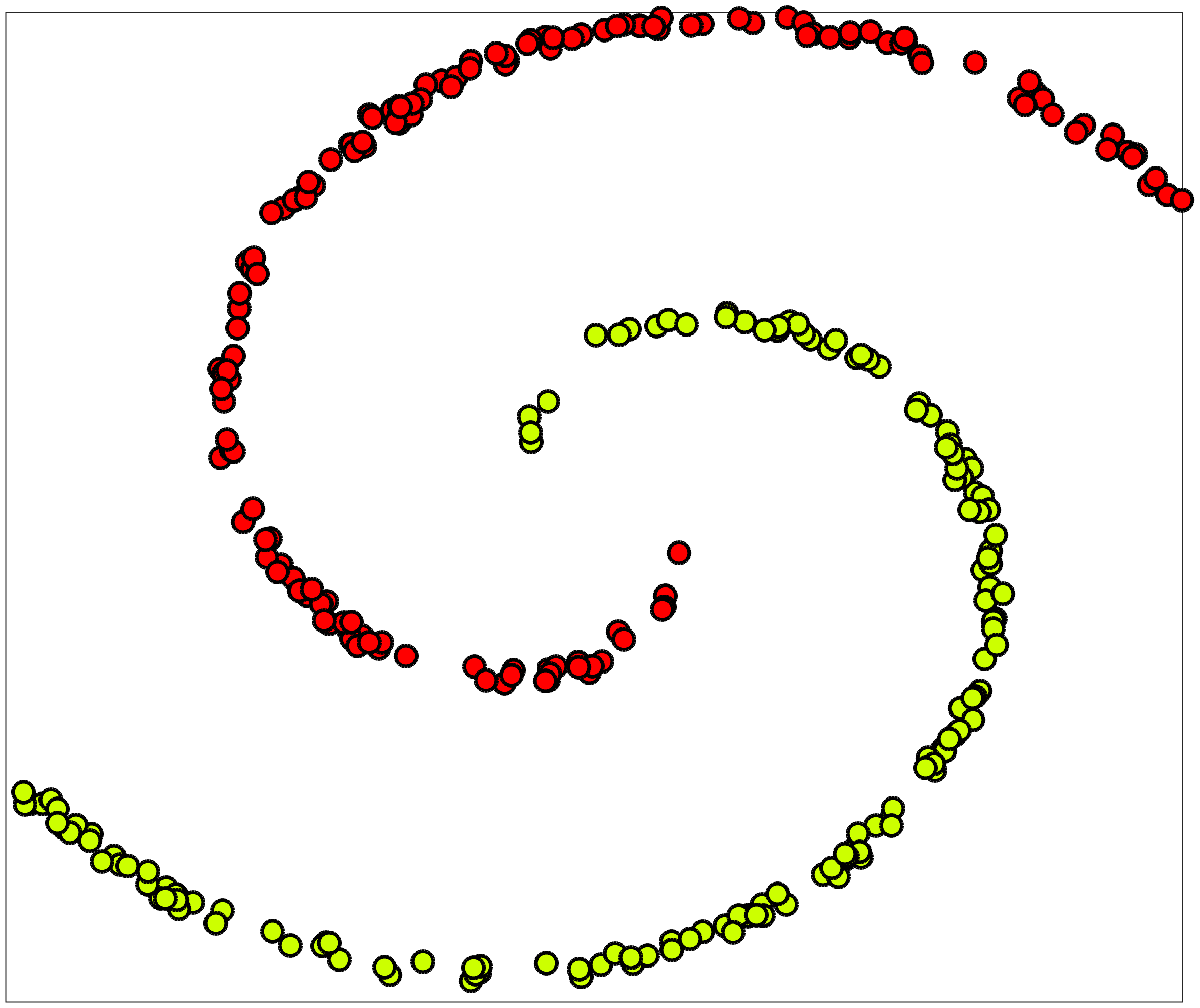}}
 \caption{Two spirals dataset. We demonstrate the proposed local scale estimation from the conditional quantiles of the similarity function. The reconstructed affinities (b-d) are used to measure the rate at which edges decay as a function of $\tau$ (e) and subsequently used to construct the stochastic graph (f). Finally, the results of spectral clustering reveal the effectiveness of the proposed approach.}
\label{fig:twospirals}
\end{figure*}

Our proposed approach begins by constructing a locally scaled affinity matrix using (\ref{eqn:localscaling}) and inferring the conditional quantiles using the approach described in Section \ref{sec:optimization}. More specifically, we train a set of auto-encoding neural networks following the architecture in Figure \ref{fig:AE} to recover the affinity matrix at different values of the quantile parameter, $\tau = \{0.1,0.2,0.3,0.4,0.5,0.6,0.7,0.8,0.9\}$. In each network, we use the locally scaled affinity matrix ($\mathbf{W}_{\ell}$) as the input and obtain the output from the decoder layer, $\mathbf{W}_{\ell}^{\tau}$, representing the affinity matrix at the $\tau^{\text{th}}$ quantile. An interesting observation is that this approach produces a unique set of neighborhood graphs monotonously parameterized by $\tau$. In other words, the graph monotonously loses edges as $\tau$ increases and depending on the data distribution, becomes disconnected at a certain quantile. While it might be interesting to analyze the clusters revealed at different quantiles of the similarity function, we believe that the rate at which the graph loses its edges directly reveals the significance of the edges in recovering the underlying structure. For example, let us consider the synthetic data in Figure \ref{fig:twospirals}(a) that contains samples from two spirals. While the original affinity matrix is significantly dense, by ignoring edges with trivial affinities, the recovered graph at the $0.1$ quantile (Figure \ref{fig:twospirals}(b)) contains only $4939$ edges. Using only a fraction of the edges, it still recovers the underlying structure effectively. As shown in Figure \ref{fig:twospirals}(e), the graph becomes increasingly sparse at higher quantiles and creates disconnected components. While intuition suggests that edges which persist at higher quantiles are crucial to recovering the underlying clusters, it is also important to note that the significance of an edge depends on the level of sparsity in the recovered graph, regardless of the quantile used. The latter observation will ensure that edges from regions with disparate densities are treated equally. Hence, we propose to construct a stochastic graph, wherein the probability of an edge is proportional to the sparsity of the graph from the highest $\tau$ at which the edge persists before disappearing.

Denoting the number of edges in the affinity matrix recovered at quantile $\tau$ as $\beta_{\tau}$, the probability of an edge, $e_{ij}$ is measured as
\begin{equation}
p(e_{ij}) = max\left(\delta, 1 - \frac{\beta_{\hat{\tau}}}{\beta_{0.1}}\right),
\label{eqn:prob}
\end{equation}where $\hat{\tau}$ corresponds to the highest quantile at which $e_{ij}$ persists and $\delta$ is the minimal probability assigned to all edges in the graph at $\tau=0.1$ (in all our experiments, $\delta = 0.4$). Figure \ref{fig:twospirals}(f) illustrates the estimated stochastic graph for the two spirals dataset. 

\begin{figure}[t]
\centering
  \includegraphics[width=.99\linewidth]{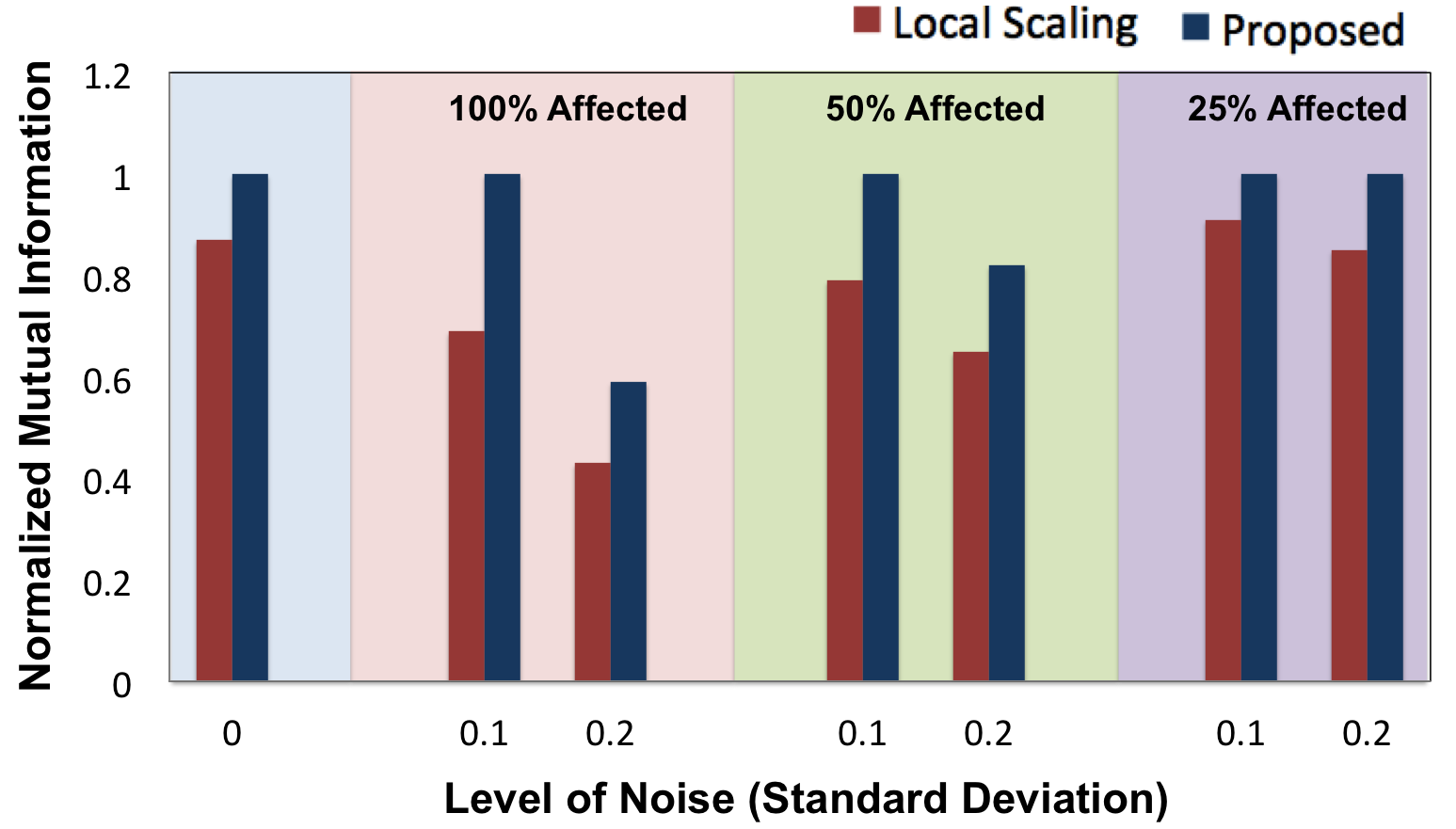}
 \caption{Impact of noise on clustering performance of local scaling in \ref{eqn:localscaling} and the proposed approach with the twospirals dataset in Figure \ref{fig:twospirals}(a).}
\label{fig:noise}
\end{figure}
  
  \begin{figure*}[t]
\centering
  \subfigure[Input Data]{\includegraphics[width=.23\linewidth]{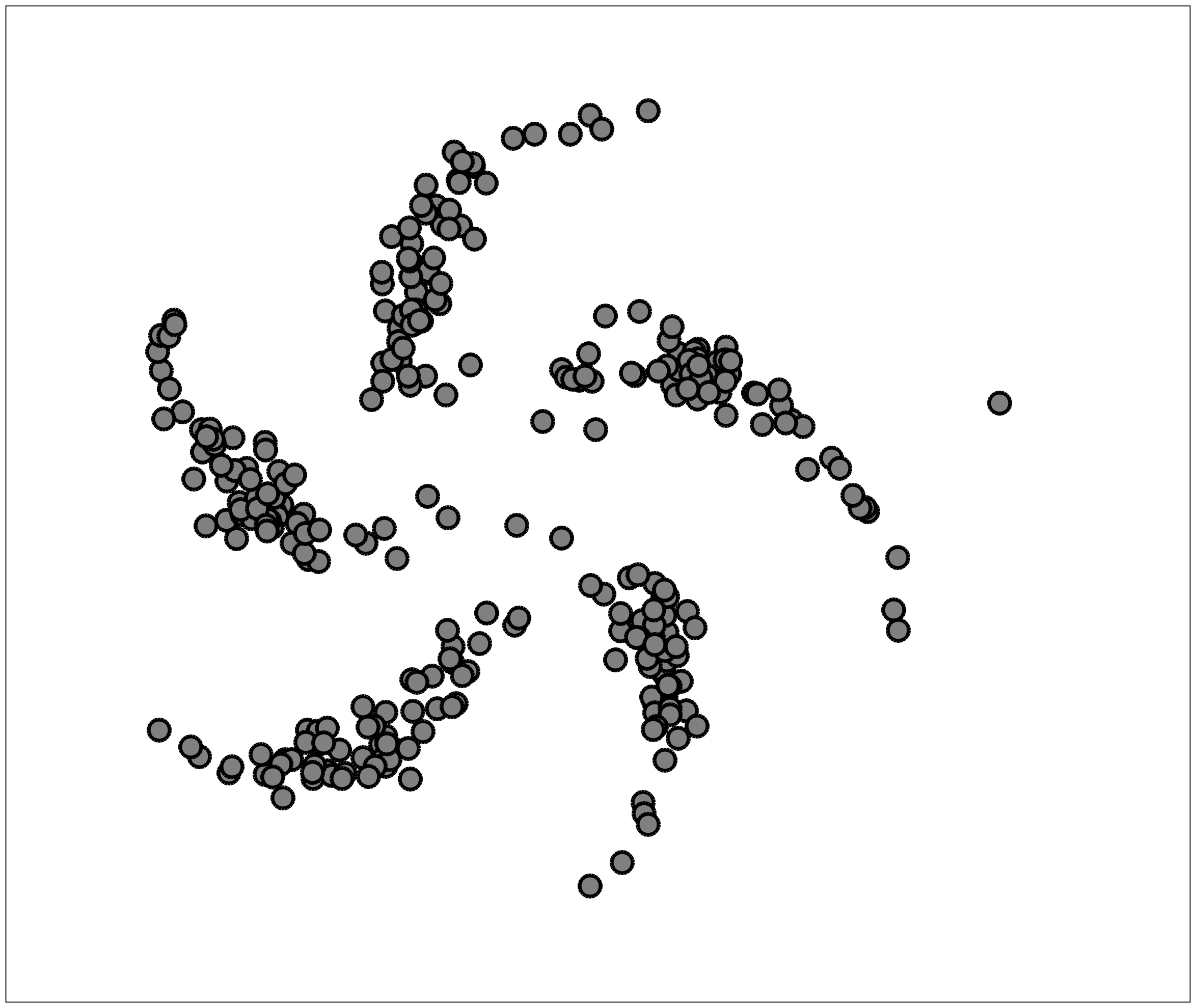}}\hfill
  \subfigure[Edge Decay]{\includegraphics[width=.23\linewidth]{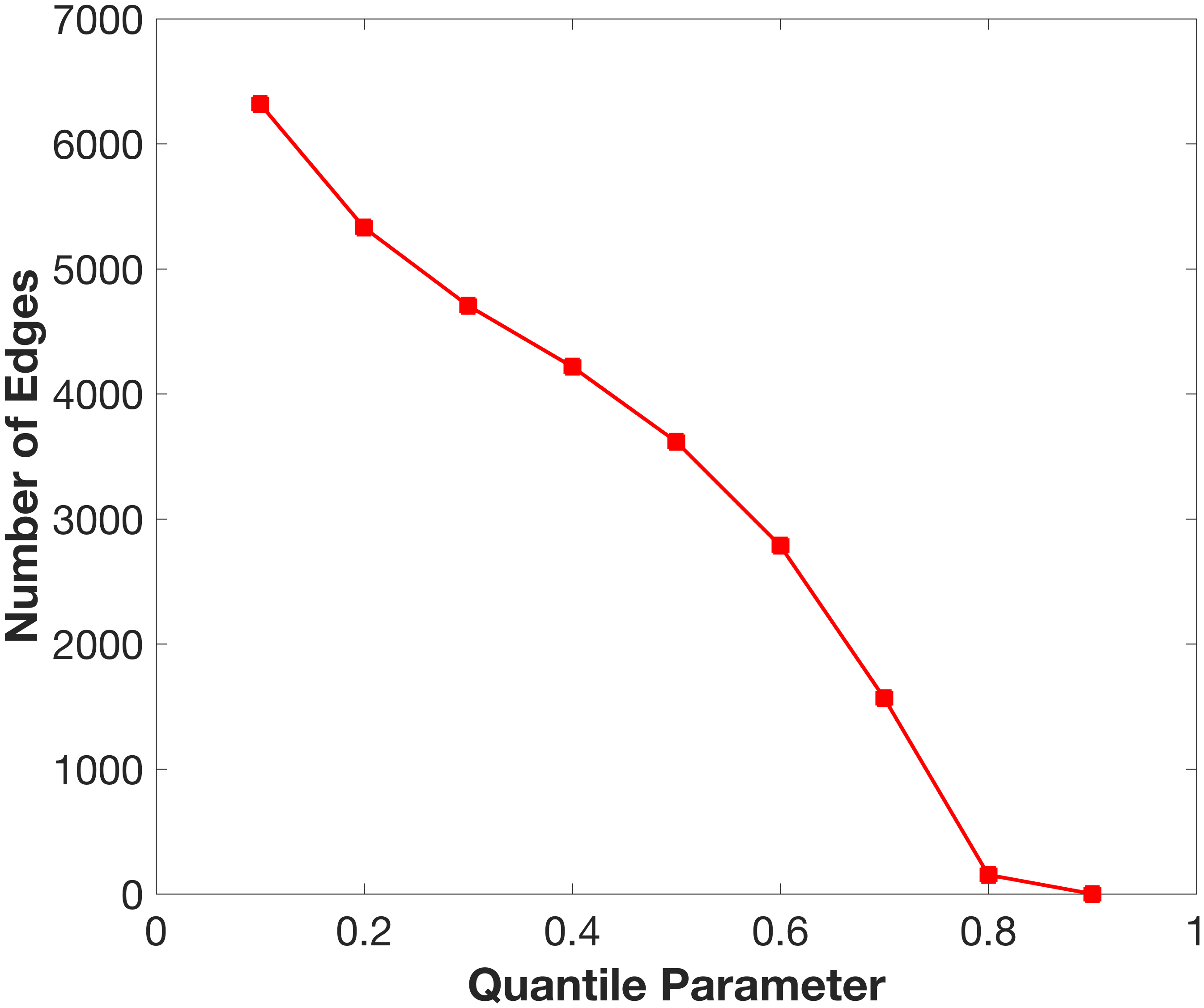}}\hfill
  \subfigure[Stochastic Graph]{\includegraphics[width=.23\linewidth]{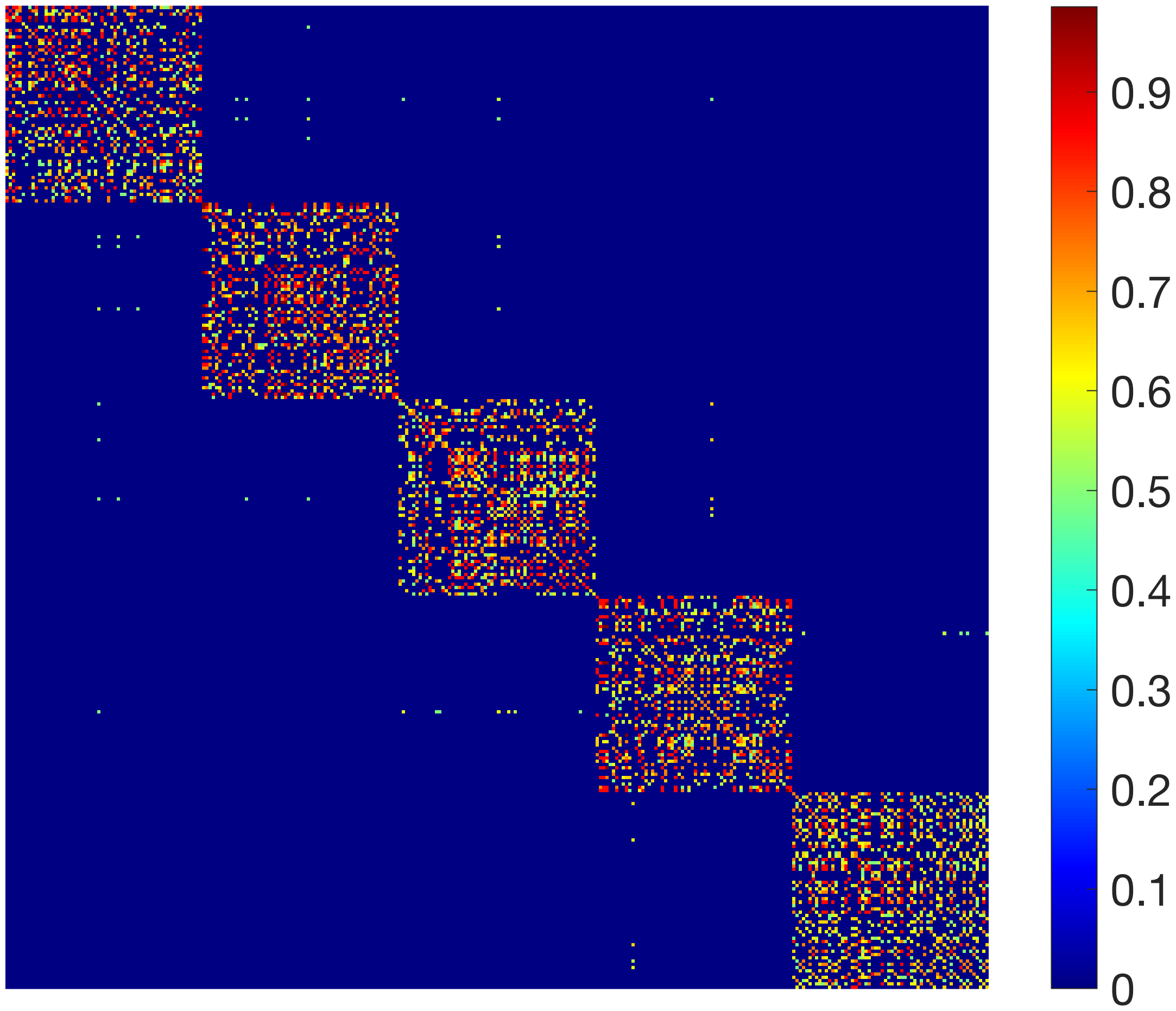}}\hfill
  \subfigure[Clustering]{\includegraphics[width=.23\linewidth]{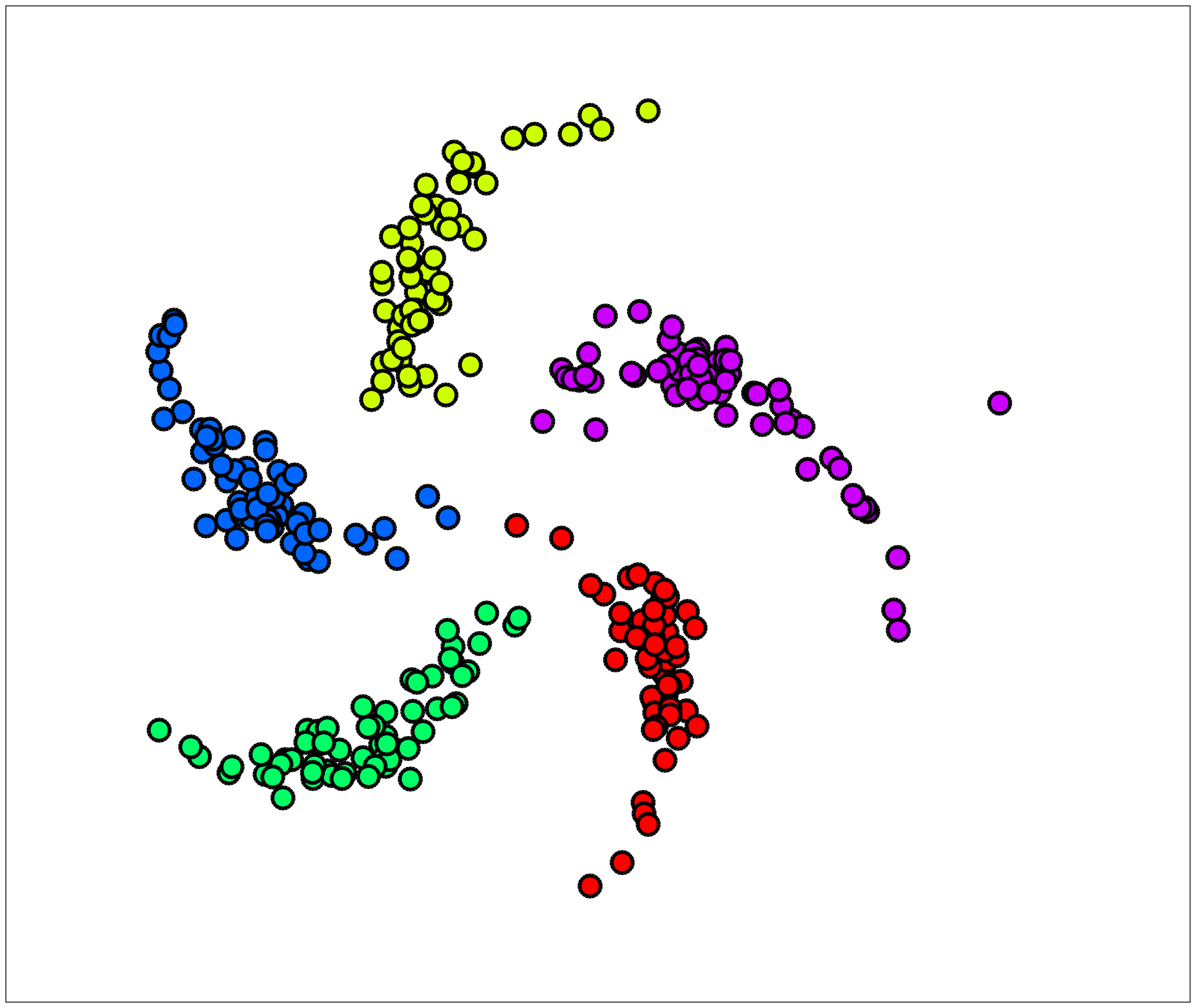}}
 \caption{Pin wheel dataset. Robust scale estimation leads to accurate recovery of the underyling clusters.}
\label{fig:pinwheel}
\end{figure*}

\subsection{Algorithm}
\label{sec:algo}
In this section, we describe the proposed algorithm for obtaining robust local scale estimates using the stochastic graphs. Existing approaches such as consensus clustering \cite{race2014flexible} directly use stochastic graphs as inputs to spectral clustering. An important downside of such approaches is that they require multiple iterations of refinement to construct robust graph affinities. In contrast, we propose to generate multiple random realizations of the neighborhood using the inferred edge probabilities and estimate the local scales in each of the realizations independently. Denoting a random realization of the stochastic graph as $\tilde{\mathbf{W}}^{(1)}$, we estimate the local scale for a sample $\mathbf{x}_i$ as
\begin{equation}
\sigma_i^{(1)} = \frac{1}{|\mathcal{N}_i|} \sum_{j \in \mathcal{N}_i} d(\mathbf{x}_i, \mathbf{x}_j),
\end{equation}where $\mathcal{N}_i$ denotes the set of neighbors of $\mathbf{x}_i$ identified using $\tilde{\mathbf{W}}^{(1)}$. The estimate of the local scale is obtained as the median of the local scales at each of the realizations, $\sigma_i = median\left([\sigma_i^{(1)}, \sigma_i^{(2)}, \cdots, \sigma_i^{(R)}]\right)$, where $R$ is the total number of independent realizations. Finally, the estimated scaled are used to construct the graph affinities as shown in (\ref{eqn:localscaling}). The steps of the algorithm are summarized in Algorithm 1. Figure \ref{fig:twospirals}(g) and (h) show the spectral clustering results obtained with the local scaling approach in (\ref{eqn:localscaling}) and the proposed approach respectively. As it can be observed, using the conditional quantiles leads to robust scale estimates and consequently the proposed approach accurately recovers the two spirals. We also analyze the sensitivity of the resulting graphs to noise in the data, by adding random Gaussian noise ($\sigma^{noise} = \{0.1,0.2\}$) to different number of randomly chosen samples ($25\%, 50\%, 100\%$). From the results in Figure \ref{fig:noise}, the proposed scaling approach is highly resilient to the noise in data in all cases. Figure \ref{fig:pinwheel} illustrates the estimated probabilities and clustering performance of the proposed method on the pinwheel dataset. Another interesting observation is that the choice of parameter $k$ in the construction of the initial affinity $\mathbf{W}_{\ell}$ does not significantly affect the performance of the proposed approach. For example, we evaluated the clustering results on the twospirals data by varying $k$ between $5$ and $15$ and found that our method consistently recovers the true clusters effectively.

\begin{table*}[t]
\label{table:resclustering}
\renewcommand{\arraystretch}{1.5}
\centering
\begin{tabular}{|c|c|c|c|c|}
\hline
\textbf{Dataset} & \textbf{Local Scaling}\cite{perona} & \textbf{Consensus Clustering}\cite{race2014flexible} & \textbf{Path-based Similarities}\cite{Chang2005}& \textbf{Proposed Approach}\\
\hline
Two spirals&0.87 &\textbf{1.0} &0.92 &\textbf{1.0}\\
Pinwheel&0.95 &0.97 &0.97 &\textbf{1.0}\\
Glass&0.37 &0.38 &0.39 & \textbf{0.42}\\
Breast Cancer&0.79 &0.82 &0.78 &\textbf{0.85}\\
Wine&0.91 &0.89 &0.91 &\textbf{0.93}\\
E-coli&0.57 &\textbf{0.63} &0.58 &\textbf{0.63}\\
Leaf& 0.66&0.73 &0.68 &\textbf{0.75}\\
Parkinson&0.31 & 0.32&0.33 &\textbf{0.37}\\                
\hline
\end{tabular}
\vspace{0.1in}
\caption{Performance comparison of spectral clustering. We compare the proposed robust affinities with baseline graph construction techniques. In each the method that produces the highest NMI is marked in bold.}
\end{table*}

\section{Experiments}
\subsection{Spectral Clustering}
In this section, we evaluate the performance of the proposed graph construction approach in spectral clustering. We have explored our approach using a number of datasets from the UCI machine learning repository \cite{UCI}. These data sets are characterized for having clusters of varying density, scale and shape---where spectral algorithms using a global scale are known to perform poorly. We evaluate the clustering performance using the Normalized Mutual Information (NMI) score, defined as
\begin{equation}
NMI(Y;\hat{Y}) = \frac{2 I(Y;\hat{Y})}{H(Y) + H(\hat{Y})},
\end{equation}where $I(Y;\hat{Y})$ is the mutual information between sets $Y$ and $\hat{Y}$, and $H(Y), H(\hat{Y})$ are the entropies of the two sets. 

For comparison, we computed the clustering performance of the local scaling approach in \cite{perona} and the path-based similarities in \cite{Chang2005}. In addition, we implemented a consensus clustering approach, which used different values for the parameter $k$ to obtain the local scale estimates and adopted the consensus inference approach in \cite{race2014flexible} to perform clustering. For the proposed approach, we fixed the number of random realizations in Algorithm 1, $R=25$. Table 1 shows the NMI obtained using the different techniques and our approach outperforms the baseline techniques in all cases. 

\subsection{Label Propagation using a Single Example}
Another interesting application of graph similarities is in propagating labels to a large set of test examples using a limited training set. In its extreme case, the problem of classification with only one labeled example per class can benefit significantly from robust graphs \cite{Chang2006}. In this section, we develop a greedy label propagation strategy based on graph similarities and evaluate the effectiveness of the proposed graph construction in this application.

\begin{algorithm}[t]
  \label{alg:prop}
  \textbf{Input:} Data $\mathbf{X}$ with the first $C$ samples labeled, graph affinity $\mathbf{W}$
  \begin{enumerate}
  \item Compute the graph laplacian using (\ref{eqn:laplacian}) and build the graph laplacian kernel $\mathbf{K} = \mathbf{L}^{\dagger}$
  \item Measure Euclidean distances on the RKHS to construct the distance matrix $\mathbf{S}$
  \end{enumerate}
  \noindent For each unlabeled example $j$, initialize $ind = j$, $iter = 0$,
  \begin{enumerate}
  \setcounter{enumi}{2}
  \item \textbf{while} $iter < maxwalk$:
  \begin{enumerate}
  \item Store the distances $\gamma^{iter}(\mathbf{x}_{ind},\mathbf{x}_i) = \min\left(S(ind,i),\gamma^{iter-1}(\mathbf{x}_{ind},\mathbf{x}_i)\right), \forall i$
  \item Determine the nearest neighbor for $\mathbf{x}_{ind}$, $t = arg \min \mathbf{S}(ind,:)$ 
  \item \textbf{if} $t \in \{1,\cdots,C\}$, \textbf{break}
  \item $ind = t, iter = iter + 1$
  \end{enumerate}
  \item Label$(\mathbf{x}_j) = arg \min_i \boldsymbol{\gamma}^{iter}$
  \end{enumerate}
  
  \caption{Perform label propagation using greedy walk on the graph laplacian kernel}
  \end{algorithm}

\subsubsection{Greedy Walk on Graph Laplacian Kernels}
Existing methods for label propagation often rely on graph-based methods with local and global consistency constraints \cite{Zhou2004}. In particular, constructing appropriate RKHS (Reproducing Kernel Hilbert Space) kernels by transforming the spectrum of the graph over labeled and unlabeled data together has been effective. Hence, we adopt the approach in \cite{Chang2006} to construct a graph laplacian kernel from weighted neighborhood graphs for label propagation. 

Given the graph affinity matrix $\mathbf{W} \in \mathbb{R}^{N \times N}$, we construct the normalized graph laplacian $\mathbf{L}$ (not to be confused with the matrix $\mathbf{L}$ in Section \ref{sec:qloss}) as follows:
\begin{equation}
\mathbf{L} = \mathbf{D}^{-\frac{1}{2}} (\mathbf{D} - \mathbf{W}) \mathbf{D}^{-\frac{1}{2}},
\label{eqn:laplacian}
\end{equation}where $\mathbf{D}$ is the degree matrix whose diagonal entries are defined as $D_{ii} = \sum_j w_{ij}$. Let $\mathcal{F}(G)$ denote the linear space of real-valued functions defined on the graph G and $\{\lambda_i,\mathbf{u}_i\}_{i = 1}^N$ denote the eigen spectrum of the corresponding laplacian $\mathbf{L}$. Now, we define a Hilbert space of functions on G, $\mathcal{H}(G) = \{\mathbf{g}|\mathbf{g}^T \mathbf{u}_i = 0, \forall i\}$, which is a linear subspace of $\mathcal{F}(G)$ orthogonal to the eigenvectors of $\mathbf{L}$ with zero eigenvalues. Similar to the analysis in \cite{Herbster2005}, we can show that the pseudo-inverse of $\mathbf{L}$ is the reproducing kernel of $\mathcal{H}(G)$. The resulting matrix $\mathbf{K} = \mathbf{L}^{\dagger}$ is referred to as the \textit{graph laplacian kernel}. 

In our problem setup, the data contains $C$ classes with one labeled example per class. Without loss of generality, we assume that the first $C$ samples in the input dataset correspond to the labeled examples and the rest are unlabeled. Note that, the distance between two samples can be obtained by measuring their Euclidean distance in the RKHS,
\begin{align}
d(\mathbf{x}_i, \mathbf{x}_j) &= \|\phi(\mathbf{x}_i) - \phi(\mathbf{x}_j)\|_2 = K_{ii} + K_{jj} - 2K_{ij}.
\label{eqn:prop}
\end{align}Using this, we build the distance matrix $\mathbf{S} \in \mathbb{R}^{N \times N}$ and employ a greedy walk scheme, shown in Algorithm 2, for propagating the labels. In all our experiments, the maximum length of the greedy walk, $maxwalk$, was fixed at $5$. The performance of this propagation strategy relies heavily on the robustness of the graph similarity $\mathbf{W}$ to noise and outliers.

\subsubsection{Results}
\begin{figure*}[t]
\centering
  \subfigure[Echocardiogram]{\includegraphics[width=.48\linewidth]{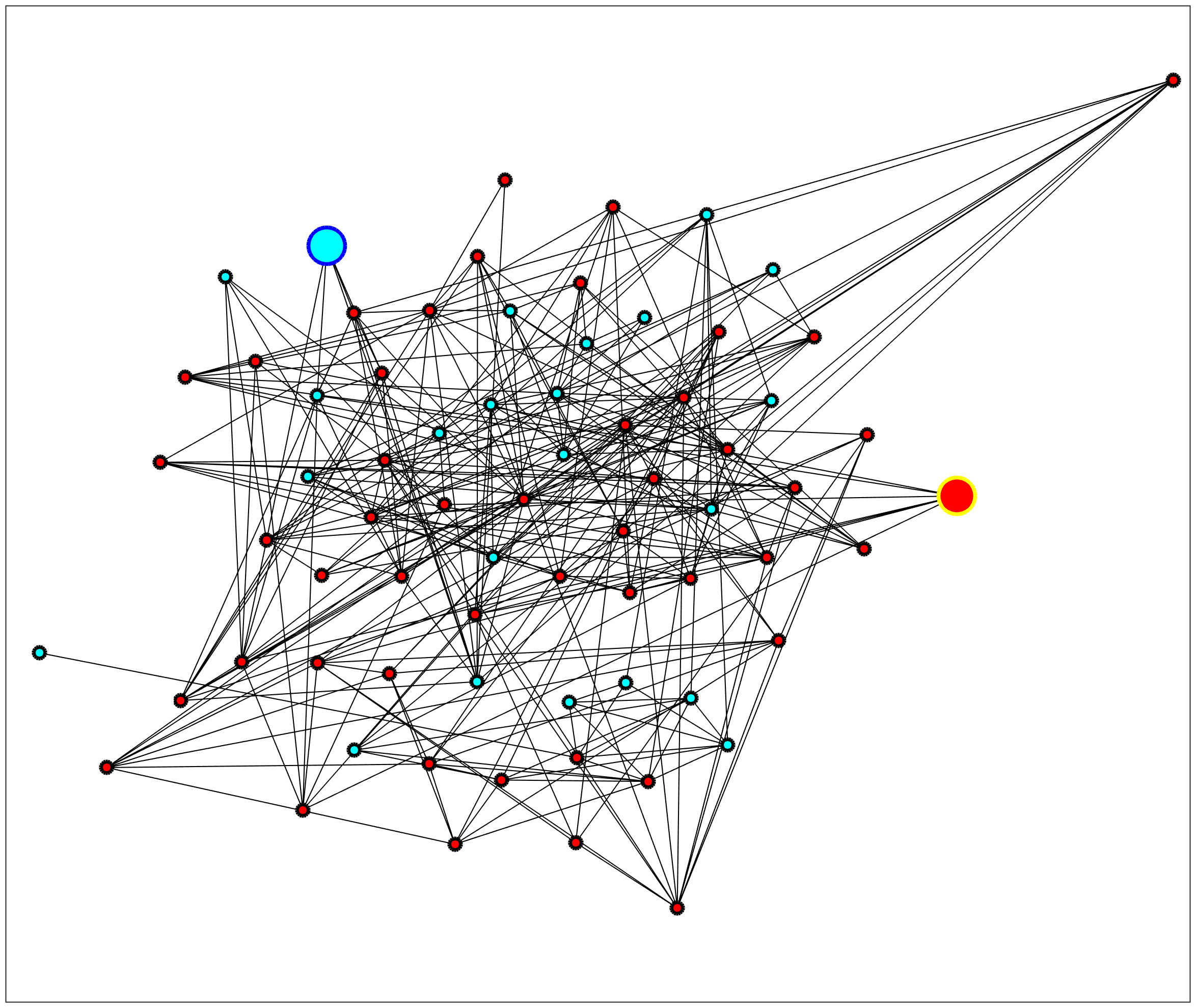}}\hfill
  \subfigure[Breast Cancer]{\includegraphics[width=.48\linewidth]{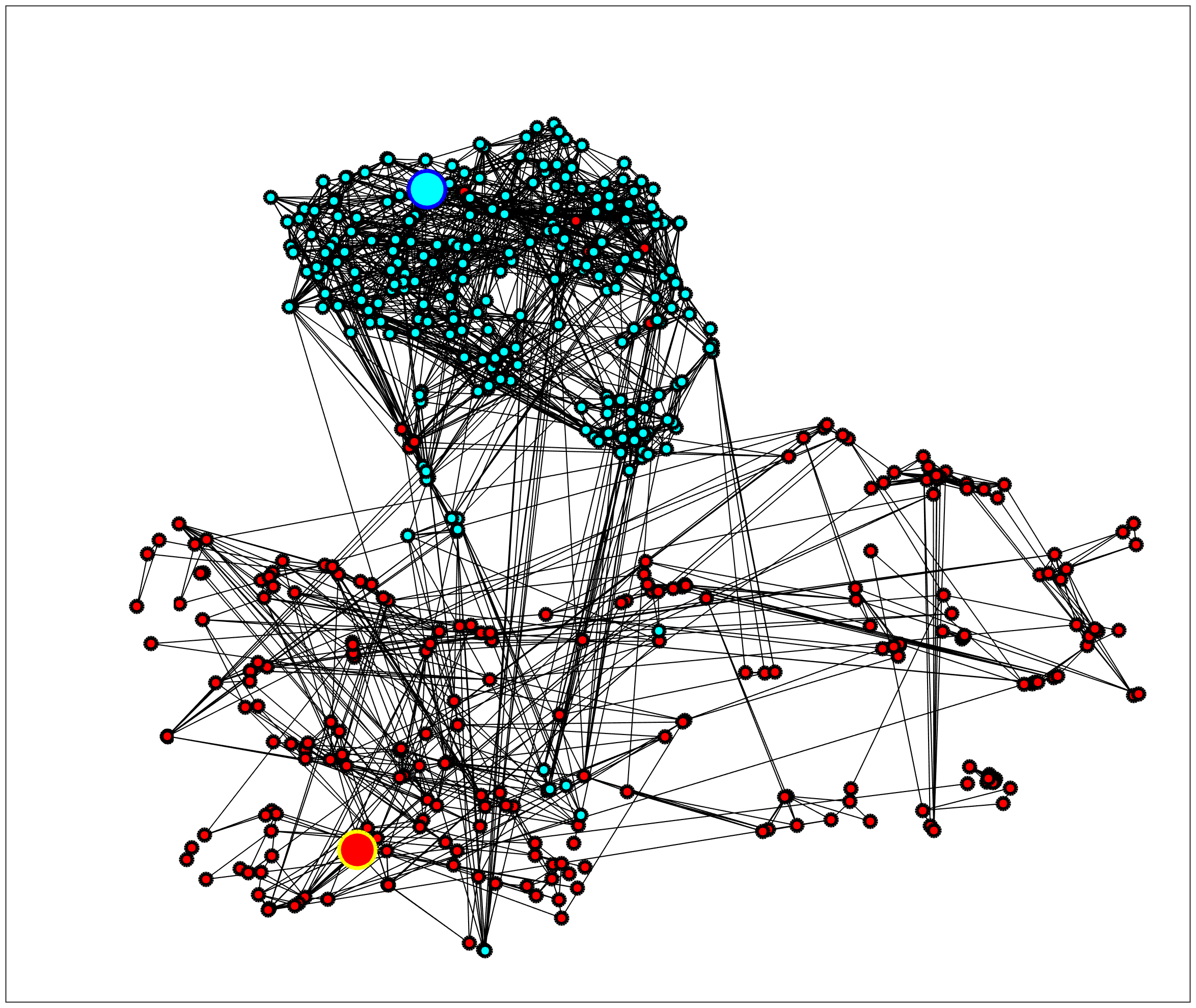}}\hfill

 \caption{Locally scaled affinities for the label propagation experiments. The two training examples are shown as the bigger circles. The $2-$D embeddings of the samples are created, using the t-SNE algorithm, for visualization.}
\label{fig:classdemo}
\end{figure*}

We evaluate the performance of different locally scaled affinity matrices in label propagation using a variety of challenging binary classification datasets from the UCI repository. In each dataset, we randomly chose one training example from each class and computed the accuracy of the label propagation scheme in Algorithm 2. We repeated the experiment for $10$ independent trials and the average classification accuracies are shown in Table 2. As in the spectral clustering experiments, we compared the proposed approach to (\ref{eqn:localscaling}) and path-based similarities \cite{Chang2005}. The effectiveness of the proposed neighborhood graphs is apparent from the improvements in the classification performance (as high as $18\%$) over the baseline methods. Figure \ref{fig:classdemo} illustrates the graphs obtained using the proposed approach for the \textit{echocardiogram} and \textit{breast cancer} datasets. 

\begin{table}[t]
\renewcommand{\arraystretch}{1.5}
\centering
\label{table:resclass}
\begin{tabular}{|c|c|c|c|}
\hline
\multirow{2}{*}{\textbf{Dataset}} & \textbf{Local} & \textbf{Path-based}& \textbf{Proposed}\\
& \textbf{Scaling \cite{perona}}& \textbf{Similarities}\cite{Chang2005}& \textbf{Approach}\\
\hline
Blood Transfusion& 76.05&	77.1	&\textbf{84.9}\\
Breast Cancer&82.7&84.9 & \textbf{90.55}\\
Echocardiogram& 70.06&73.2	&\textbf{88.49}\\
Kidney Disease&66.8&67.5&	\textbf{71.9}\\
SPECT Heart&70	&68.5&\textbf{86}\\
Thoracic Surgery&68.3	&66.2&\textbf{75.8}\\
Arcene&59&61.4&\textbf{71.3}\\               
\hline
\end{tabular}
\vspace{0.1in}
\caption{Performance of different graph similarity construction approaches in classification using a single example.}
\end{table}

\bibliographystyle{IEEEtran}
\bibliography{refs.bib}

\end{document}